\pdfoutput=1

\documentclass[11pt]{article}

\usepackage[]{acl}

\usepackage{times}
\usepackage{latexsym}

\usepackage[T1]{fontenc}

\usepackage[utf8]{inputenc}

\usepackage{microtype}

\usepackage{url}
\usepackage{graphicx}
\usepackage{array,multirow}
\usepackage{url}
\usepackage{bm}
\usepackage{soul}
\usepackage{amsmath}
\usepackage[normalem]{ulem}
\usepackage{makecell}
\usepackage{upgreek}
\usepackage[ruled, lined, linesnumbered, commentsnumbered, longend]{algorithm2e}
\usepackage{algpseudocode}
\usepackage{caption}
\usepackage{subcaption}
\usepackage{multirow}

\title{$\mathbf{VoteTRANS}$: Detecting Adversarial Text without Training\\by Voting on Hard Labels of Transformations}

\author{Hoang-Quoc Nguyen-Son$^\textsuperscript{1}$, Seira Hidano$^\textsuperscript{1}$, Kazuhide Fukushima$^\textsuperscript{1}$, \\ {\bf Shinsaku Kiyomoto$^\textsuperscript{1}$, \and Isao Echizen$^\textsuperscript{2}$ } \\
  $^\textsuperscript{1}$KDDI Research, Inc., Japan \\
  $^\textsuperscript{2}$National Institute of Informatics, Japan \\
  $^\textsuperscript{1}${\tt \{xso-guen,se-hidano,ka-fukushima,sh-kiyomoto\}@kddi.com} 
  \\
  $^\textsuperscript{2}${\tt iechizen@nii.ac.jp} 
  }
  
\begin{document}
\maketitle
\begin{abstract}
Adversarial attacks reveal serious flaws in deep learning models.
More dangerously, these attacks preserve the original meaning and escape human recognition.
Existing methods for detecting these attacks need to be trained using original/adversarial data.
In this paper, we propose detection without training by voting on hard labels from predictions of transformations, namely, $\mathrm{VoteTRANS}$.
Specifically, $\mathrm{VoteTRANS}$ detects adversarial text by comparing the hard labels of input text and its transformation.
The evaluation demonstrates that $\mathrm{VoteTRANS}$ effectively detects adversarial text across various state-of-the-art attacks, models, and datasets.

\end{abstract}

\section{Introduction}
\label{sec:introduction}
Deep learning models are sensitive to changes in input text from an adversarial attack.
Even a slight change enormously impacts the prediction of models.
More dangerously, these changes still preserve the input meaning, so attacks remain unrecognized by humans.
This vulnerability has negatively affected the reputation of deep learning models.

In contrast to adversarial text defense, fewer works have been proposed to detect adversarial texts.
Previous works detected such texts via perturbed word identification~\citep{zhou2019learning,mozes2021frequency}, synonyms~\citep{wang2021randomized}, density~\cite{yoo2022detection}, attention~\cite{biju2022input}, PCA~\cite{raina2022residue},  transformer~\cite{wang2022distinguishing}, and word importance~\citep{mosca2022suspicious}.
Since existing works need original/adversarial data to train detectors, they are sensitive to new adversarial attacks.

\begin{figure*}[!ht]
\centering
\includegraphics[]{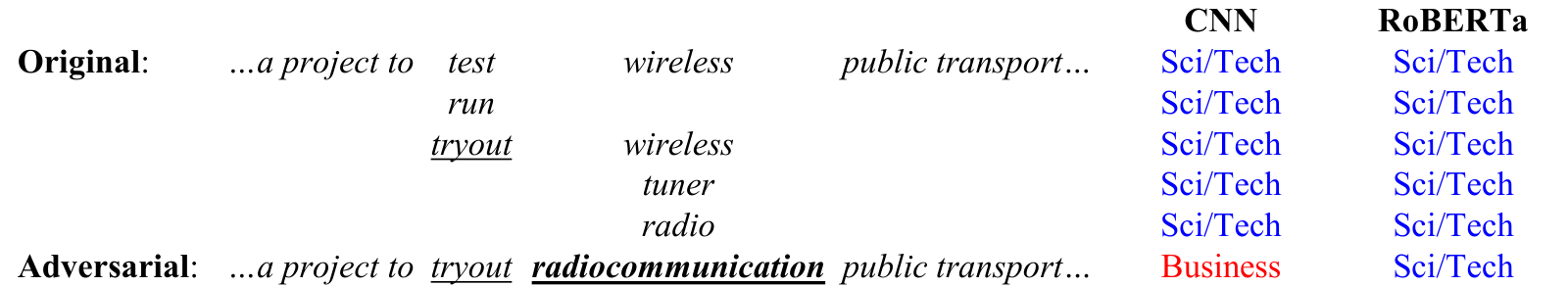}
\caption{A process generates adversarial text by synonym-based transformation targeting a CNN model. During this process, only the final adversarial text fools the CNN, while other texts are still correctly predicted by the CNN and RoBERTa.}
\label{fig:example}
\end{figure*}

\textbf{Motivation:} Adversarial text must satisfy two criteria: the text must (1) change the prediction of a target model while (2) preserving the original meaning.
Few texts can comply with both criteria.
For example, we randomly selected original text from AG News and used a probability-weighted word saliency (\textit{PWWS}) attack~\citep{ren2019generating} to generate adversarial text (Figure~\ref{fig:example}).
\textit{PWWS} replaces original words to fool a target model (CNN).
During this generation process, only the final text fooled the target CNN, while other texts were still correctly predicted by the target CNN and another model, such as RoBERTa. 
We find the same trend for 
other AG News texts and IMDB movie reviews as shown in Appendix~\ref{section:appendix:max_rate}.

\textbf{Contributions:}
We propose a simple detector by voting on hard labels of transformations  ($\mathrm{VoteTRANS}$). 
In particular, we generate a transformation set for each word in the input text.
We then compare the original hard label from the input text and the majority vote from each transformation set.
If we find any difference in the comparison, the adversarial text is identified.
In summary, our contributions are listed as follows:

\begin{itemize}
    \item To the best of our knowledge, $\mathrm{VoteTRANS}$ is the first model to detect adversarial text from various attacks without training.
    Moreover, we do not modify a target model and only use the target as a black-box setting for prediction.
    $\mathrm{VoteTRANS}$ can thus be applied to a wide range of various models.
    \item
    Experiments on various attacks, models, and datasets demonstrate that $\mathrm{VoteTRANS}$ outperforms state-of-the-art detectors.
    \item
    $\mathrm{VoteTRANS}$ can run with all seventeen current attacks related to text classification from \mbox{TextAttack} framework~\cite{morris2020textattack}. 
    $\mathrm{VoteTRANS}$ is also automatically compatible with future attacks from this framework without changing its source code\footnote{$\mathrm{VoteTRANS}$ is available at \url{https://github.com/quocnsh/VoteTRANS}}.
\end{itemize}

\section{Related Work}

\subsection{Adversarial Attack}

Many adversarial attacks have emerged since 2018 and have been supported by TextAttack~\cite{morris2020textattack}.
We categorize all seventeen attacks from \mbox{TextAttack} related to text classification by their levels.

In word level, \textit{Alzantot}~\cite{alzantot2018generating}, \textit{Faster Alzantot}~\cite{jia2019certified}, and an improved generic algorithm (\textit{IGA})\cite{wang2021natural} ran a generic algorithm to generate adversarial text.
\textit{PWWS}~\cite{ren2019generating} and \textit{TextFooler}~\cite{jin2020bert} transformed a text with synonyms from a fixed lexical database (WordNet) and word embedding, respectively.
\newcite{zang2020word} applied particle swarm optimization (\textit{PSO}) to  synonyms from a big database (HowNet).
\textit{Kuleshov}~\cite{kuleshov2018adversarial} and \textit{A2T}~\cite{yoo2021towards} measured the similarity with \mbox{GPT-2} and DistillBERT, respectively.
\textit{BERT-Attack}~\cite{li2020bert} extracted the synonyms from a masked language model.
\textit{BAE}~\cite{garg2020bae} used both word insertion and replacement with the masked model.
\textit{CLARE}~\cite{li2021contextualized} extended this model by merging two consecutive words.
\textit{Checklist}~\cite{ribeiro2020beyond} verified the model consistency by changing an input word into a neutral entity (e.g., location, name, or number).
\textit{Input-Reduction}~\citep{feng2018pathologies} removes the word of lowest importance until the target model changes the prediction.

In character and hybrid levels, \textit{HotFlip}~\cite{ebrahimi2018hotflip} accessed the gradient of a target model to manipulate the loss change.
\textit{DeepWordBug}~\cite{gao2018black} transformed words using four-character operations including swapping, substitution, deletion, and insertion.
\textit{Pruthi}~\cite{pruthi2019combating} added a QWERTY keyboard operator.
\textit{TextBugger}~\cite{li2019textbugger} combined character and word operators.

\subsection{Adversarial Detection}

\citet{zhou2019learning} trained a BERT model to identify the perturbed words.
\citet{mozes2021frequency} focused on low-frequency words that were likely changed by an attacker.
\citet{wang2021randomized}  detected the adversarial text via its synonyms.
\citet{yoo2022detection} and \citet{biju2022input} distinguished adversarial text with original text based on density estimation and attention input, respectively.
\citet{raina2022residue} showed that adversarial text induces residues (larger components than original text) in PCA eigenvector.
\citet{wang2022distinguishing} finetuned a transformer model
on compliant and adversarial text, 
which is not required to fool a target model. 
The change in the requirement efficiently defends against a wide range of adversarial attacks.

\citet{mosca2022suspicious} extracted the word importance as a feature to classify original and adversarial text.

Compared with $\mathrm{VoteTRANS}$, previous detectors need adversarial or/and original data to train their models
or optimizes their models on training data
to satisfy some requirements (such as FPR as in \citet{yoo2022detection}).
These methods often more suitable for word-based than character-based attacks. 
On the other hand, $\mathrm{VoteTRANS}$ can be applied to various kinds of attacks at both word-level as in \textit{PWWS} and character-level as in \textit{TextBugger} as well as other modifications, such as deletion as in \textit{Input-Reduction} or \textit{CLARE} and insertion as in \textit{BAE}.
For example, \textit{CLARE} deletes a word by merging it with a nearby word. 
When applying the same merging operator on each input word, $\mathrm{VoteTRANS}$ will change the attacked words and observe the change in target predictions to detect the attacked text.

\begin{figure*}[!ht]
\centering
\includegraphics[]{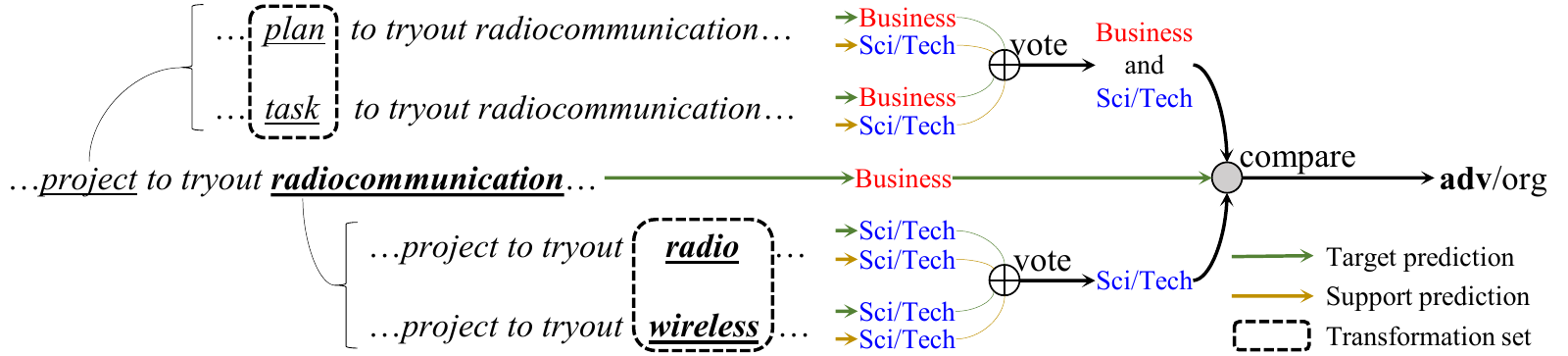}
\caption{Given input text, we replace each word  (e.g., ``\underline{\textit{project}}'' and ``\textbf{\underline{\textit{radiocommunication}}}'') with the corresponding transformation set.
The modified text is then predicted by a target and support models as hard labels.
These labels are voted and compared with the original label to determine whether the input text is adversarial or original.}
\label{fig:VoteTRANS}
\end{figure*}

\section{Voting on the Hard Labels of Transformations ($\mathbf{VoteTRANS}$)}

\textbf{Problem statement:}
We follow notations from \textit{TextFooler} paper~\cite{jin2020bert}. 
Given $N$ texts, $\mathcal{X} = \{X_1, X_2,\dots, X_N\} $ corresponding to $N$ labels $\mathcal{Y}=\{Y_1, Y_2,\dots, Y_N$\}, a target model $F: \mathcal{X} \to \mathcal{Y}$ maps the input space $\mathcal{X}$ to the label space $\mathcal{Y}$.
Adversarial text $X_{\mathrm{adv}}$ generated from input $X \in \mathcal{X}$ should comply with the following constraints:
\begin{equation}
\label{equ:constrainsts}
    F(X_{\mathrm{adv}}) \neq F(X), \text{and }  {\mathrm{Sim}}(X_{\mathrm{adv}}, X) \geq \epsilon,
\end{equation}
where ${\mathrm{Sim}}(X_{\mathrm{adv}}, X)$ measures the similarity between adversarial text $X_{\mathrm{adv}}$ and original text $X$; $\epsilon$ is the minimum similarity.

Our objective is to determine whether input text $X$ is adversarial or original text.
The process of $\mathrm{VoteTRANS}$ is depicted in Figure~\ref{fig:VoteTRANS}, and the overall algorithm is summarized in Algorithm~\ref{alg:proposal}.

\newcommand\mycommfont[1]{\small\ttfamily#1}
\SetCommentSty{mycommfont}

\begin{algorithm*}[!ht]
\caption{Adversarial text detection by $\mathrm{VoteTRANS}$.} 
\label{alg:proposal}

\SetKwInOut{KwIn}{Input}
\SetKwInOut{KwOut}{Output}

\KwIn{Input~text~$X = \{w_1, w_2\cdots\}$; Target~model~$F$; Auxiliary~attack~$A$;
    Optional:~ \{Word~ratio~$\alpha$ (100\% as default); Support~models~$F^{\mathrm{sup}} = \{F^{\mathrm{sup}}_1, F^{\mathrm{sup}}_2\cdots\}$ (an empty as default)\}
        }
\KwOut{Adversarial text detection (\textit{True}/\textit{False})}

$Y \leftarrow F(X)$

\For{ each word $w_i$ in $X$}{
    calculate importance score $I_{w_i}$ using $A$
}

Create a list $W$ of all words $w_i \in X$ by sorting in descending order of the importance score $I_{w_i}$

$W^* \leftarrow $ obtains the top words from $W$ with ratio $\alpha$

\For{each word $w_j$ in $W^*$}{

    Transformation label list $Y^{\mathrm{trans}} = \{\}$

    Create transformation set $W^{\mathrm{trans}}$ from $w_j$ using $A$

    \For{each word $w_{k}^{\mathrm{trans}}$ in $W^{\mathrm{trans}}$}{
    
        $X'=$ replace $w_j$ with $w_{k}^{\mathrm{trans}}$ in $X$
        
        \If{checking $({\mathrm{Sim}}(X', X) \geq \epsilon)$ is satisfied by using $A$}{
        
            Add $F(X')$ to list $Y^{\mathrm{trans}}$
            
            \For{each $F^{\mathrm{sup}}_{l}$ in $F^{\mathrm{sup}}$}{
            
                Add $F^{\mathrm{sup}}_{l} (X')$ to list $Y^{\mathrm{trans}}$
            }
        }
    }
    $Y'\leftarrow$ majority vote on $Y^{\mathrm{trans}}$ 
    
    \If{$(Y \not\in Y') or (Y \in Y'$ and $Y'$ \text{has more than one majority class}$)$}{        
        \KwRet{True} \Comment{adversarial text}
    }
}

\KwRet{False} \Comment{original text}

\end{algorithm*}

\textbf{Model detail:}
To process input text $X$, we use a target model $F$ and an auxiliary attack $A$ (e.g., \textit{PWWS}).
An optional word ratio $\alpha$ can be used to speed up the detection process.
Moreover, a support model list $F^{\mathrm{sup}}$ is used to improve the performance.
The support models and the target model should solve the same task such as sentiment analysis. 

First, we create a list $W$ including input words that are sorted by their importance using word importance score estimation in the auxiliary attack $A$ (lines 2-5).
For example, \textit{PWWS} calculates the change in the predictions before and after replacing a certain word with an unknown word to estimate word importance.
The impact of word importance is shown in Appendix~\ref{sec:appendix:word_importance}.

Second, we select the top words $W^*$ (line 6) with the word ratio threshold $\alpha$ (100\% by default).
Since $\mathrm{VoteTRANS}$ maintains the performance with $\alpha = 100\%$, it remarkably improves the run time with a smaller $\alpha$, with a slight change in performance, as shown in Figures~\ref{fig:threshold_AG} and \ref{fig:threshold_IMDB} in the experimental results.

Third, we obtain a transformed set $W^{\mathrm{trans}}$ for each word $w_j$ in $W^*$ (line 9) using word transformation in $A$.
For example, \textit{PWWS} uses synonyms from WordNet to form the transformed set.

Fourth, we replace each transformed word in $W^{\mathrm{trans}}$ for the corresponding $w_j$ (line 11) in $X$ and checks the second constraint in Equation~\ref{equ:constrainsts} (line 12).
The $\mathrm{Sim}(\cdot)$ and $\epsilon$ are provided by $A$.
This step is mainly used to preserve the original meaning of the input in the transformed text.
For example, \textit{PWWS} uses stop word checking for this step.

Fifth, we construct $Y^{\mathrm{trans}}$ containing the label predictions for the valid $X'$ from target model $F$ and support model list $F^{\mathrm{sup}}$ (lines 13-16).
We then use the majority vote on $Y^{\mathrm{trans}}$ to obtain the top majority classes $Y'$ with the highest occurrence in $Y^{\mathrm{trans}}$.

Finally, we check whether the input text $X$ is adversarial or original based on the input label $Y$ and majority set $Y'$.
If $Y$ does not belong to $Y'$ or $Y$ belongs to $Y'$, which contains more than one majority class, we decide that $X$ is adversarial text (lines 20-22).
If we cannot decide if $X$ is adversarial text after checking all words in $W^*$, $X$ is considered original text.

\section{Evaluation}

\subsection{Comparison}

We follow recent work~\citep{mosca2022suspicious} to conduct the same experiments including attacks, datasets, models, parameter settings, evaluation metrics, number of training/testing samples, etc.
In particular, evaluation was performed with adversarial text generated by four attacks\footnote{All attacks ran with the default settings from the \mbox{TextAttack} framework.} including \textit{PWWS}, \textit{TextFooler}, \textit{IGA}, and \textit{BAE}.
These attacks targeted four common models\footnote{All pretrained models were reused from the \mbox{TextAttack} framework.} (LSTM, CNN, DistilBERT, and BERT) on IMDB (235.72 words/text), Yelp Polarity (YELP) (135.21 words/text), AG News (38.78 words/text) and Rotten Tomatoes movie reviews (RTMR) (18.65 words/text).

We compared $\mathrm{VoteTRANS}$ with $\mathrm{FGWS}$~\citep{mozes2021frequency} and  $\mathrm{WDR}$~\citep{mosca2022suspicious}.
$\mathrm{FGWS}$ is claimed as a state-of-the-art in all existing detection papers that we know; $\mathrm{WDR}$ is one of the most recent published work.
Both $\mathrm{FGWS}$ and $\mathrm{WDR}$ need to be trained on the adversarial text generated from a specific configuration (model, data, \textit{attack}).
Table~\ref{tab:comparison} shows the results on two configurations: (DistilBERT, IMDB, \textit{PWWS}) with 3000 training samples (Table~\ref{tab:comparision_IMDB})\footnote{Since $\mathrm{VoteTRANS}$ does not need to train, it produce the same performance with any data-splits of the training configuration. 
We average the performance on 30 different data-splits of $\mathrm{FGWS}$ and $\mathrm{WDR}$ and ignore their variance scores for simplification.} and (DistilBERT, AG News, \textit{PWWS}) with 2400 training samples (Table~\ref{tab:comparision_AG}).
We generated adversarial text from testing sets and put them aside with corresponding original text to form balanced samples.
However, the two configurations (DistillBERT, RTMR, \textit{IGA}) and (DistillBERT, AG News, \textit{IGA}) had only 480 and 446 balanced samples, respectively.
We thus chose 500 balanced samples as testing data for other configurations.

\begin{table*}[!ht]
    \setlength\tabcolsep{2.0pt} 
    \begin{subtable}[]{\textwidth}
        \centering
        \begin{tabular}{l l l | c c | c c | c c | c c}
             \multicolumn{3}{ c }{\textbf{Configuration}} & \multicolumn{2}{| c }{$\mathbf{FGWS}$} & \multicolumn{2}{| c }{$\mathbf{WDR}$} & \multicolumn{2}{| c }{$\mathbf{VoteTRANS_{same}}$} & \multicolumn{2}{| c }{$\mathbf{VoteTRANS_{diff}}$}  \\

             \hline
             \textbf{Model} & \textbf{Data} & \textbf{\textit{Attack}} & \textbf{F1} & \textbf{Recall} & \textbf{F1} & \textbf{Recall} & \textbf{F1} & \textbf{Recall} & \textbf{F1} & \textbf{Recall} \\  
             \hline
             \hline
             DistilBERT & IMDB & \textit{PWWS} & 89.5 & 82.7 & 92.1 & 94.2 &  \textbf{96.9}  & \textbf{98.4}  & - &- \\
             LSTM & IMDB & \textit{PWWS} & 80.0 & 69.6 & 84.1 & 86.8 &  \textbf{94.8} & \textbf{97.6} & - & - \\
             CNN & IMDB & \textit{PWWS} & 86.3 & 79.6  & 84.3 & 90.0 & \textbf{95.4} & \textbf{98.8} & - & - \\
             BERT & IMDB & \textit{PWWS} & 89.8 & 82.7 & 92.4 & 92.5 & \textbf{97.4} & \textbf{98.4} & - & - \\
             DistilBERT & AG News & \textit{PWWS} & 89.5 & 84.6 & 93.1 & 96.1 & \textbf{95.3} & \textbf{97.6} & - & - \\
             DistilBERT & IMDB & \textit{TextFooler} & 86.0 & 77.6 & 94.2 & 97.3 & \textbf{97.8} & 99.6 & 97.7 & \textbf{100.0} \\
             DistilBERT & IMDB & \textit{IGA} & 83.8 & 74.8 & 88.5 & 95.5 & 95.8 & \textbf{99.2}  & \textbf{95.9} & \textbf{99.2}\\
             BERT & YELP & \textit{PWWS} & 91.2 & 85.6 & 89.4 & 85.3 & \textbf{97.4} & \textbf{98.4}  & - & -\\
             BERT & YELP & \textit{TextFooler} & 90.5 & 84.2 & 95.9 & 97.5 & \textbf{97.4} & \textbf{98.8} & 97.0 & 98.0\\
             \hline
             \hline
             DistilBERT & RTMR & \textit{PWWS} & 78.9 & 67.8 & 74.1 & 85.1 & \textbf{83.8} & \textbf{88.0} & - & - \\
             DistilBERT & RTMR & \textit{IGA} & 68.1 & 55.2 & 70.4 & 90.2 & \textbf{86.9} & \textbf{90.4} & 80.5 & 82.4 \\
             \hline
             \hline
             DistilBERT & IMDB & \textit{BAE} & 65.6 & 50.2 & 88.0 & 96.3 & \textbf{97.7} & \textbf{100.0} & 96.3 & 99.2 \\
             DistilBERT & AG News & \textit{BAE} & 55.8 & 44.0 & 81.0 & \textbf{95.4} & \textbf{85.8} & 93.2 & 85.3 & 92.8 \\
             DistilBERT & RTMR & \textit{BAE} & 29.4 & 18.5 & 68.5 & \textbf{82.2} & \textbf{79.1} & 80.8 & 65.5 & 60.8 \\
             
             \hline
             \multicolumn{3}{ c }{Overall average} & 77.5 & 68.4 & 85.4 & 91.7 & \textbf{93.0} & \textbf{95.7} & - & - 
    
        \end{tabular}
        \caption{$\mathrm{FGWS}$ and $\mathrm{WDR}$ trained on configuration (DistilBERT, IMDB, \textit{PWWS}).}
        \label{tab:comparision_IMDB}
     \end{subtable}

    \begin{subtable}[]{\textwidth}
        \centering
        \begin{tabular}{l l l | c c | c c | c c | c c}
             \multicolumn{3}{ c }{\textbf{Configuration}} & \multicolumn{2}{| c }{$\mathbf{FGWS}$} & \multicolumn{2}{| c }{$\mathbf{WDR}$} & \multicolumn{2}{| c }{$\mathbf{VoteTRANS_{same}}$} & \multicolumn{2}{| c }{$\mathbf{VoteTRANS_{diff}}$}  \\
             \hline
             \textbf{Model} & \textbf{Data} & \textbf{\textit{Attack}} & \textbf{F1} & \textbf{Recall} & \textbf{F1} & \textbf{Recall} & \textbf{F1} & \textbf{Recall} & \textbf{F1} & \textbf{Recall} \\  
             \hline
             \hline
             DistilBERT & AG News & \textit{PWWS} & 89.5 & 84.6 & 93.6 & 94.8 & \textbf{95.3} & \textbf{97.6}  & - & - \\
             LSTM & AG News & \textit{PWWS} & 88.9 & 84.9 & 94.0 & 94.2 & \textbf{95.5} & \textbf{98.4}  & - & - \\
             CNN & AG News & \textit{PWWS} & 90.6 & 87.6 & 91.1 & 91.2 & \textbf{96.5} & \textbf{99.2}  & - & - \\
             BERT & AG News & \textit{PWWS} & 88.7 & 83.2 & 92.5 & 93.0 & \textbf{94.6} & \textbf{98.0}  & - & - \\
             DistilBERT & IMDB & \textit{PWWS} & 89.5 & 82.7 & 91.4 & 93.0 & \textbf{96.9} & \textbf{98.4}  & - & - \\
             DistilBERT & AG News & \textit{TextFooler} & 87.0 & 79.4 & 95.7 & 97.3 & \textbf{96.7} & \textbf{98.4}  & 95.7 &  98.0\\
             DistilBERT & AG News & \textit{IGA} & 68.6 & 58.3 & 86.7 & 93.6 & \textbf{96.5} & \textbf{99.2}  & 93.2 & 95.6 \\
             BERT & YELP & \textit{PWWS} & 91.2 & 85.6 & 86.2 & 77.2 & \textbf{97.4} & \textbf{98.4} & - & - \\
             BERT & YELP & \textit{TextFooler} & 90.5 & 84.2 & 95.4 & 94.7 & \textbf{97.4} & \textbf{98.8} & 97.0 & 98.0\\
             \hline
             \hline
             DistilBERT & RTMR & \textit{PWWS} & 78.9 & 67.8 & 75.8 & 78.5 & \textbf{83.8} & \textbf{88.0} & - & - \\
             DistilBERT & RTMR & \textit{IGA} & 68.1 & 55.2 & 73.7 & 85.4 & \textbf{86.9} & \textbf{90.4}  & 80.5 & 82.4 \\
             \hline
             \hline
             DistilBERT & IMDB & \textit{BAE} & 65.6 & 55.2 & 88.1 & 97.0 & \textbf{97.7} & \textbf{100.0} & 96.3 & 99.2 \\
             DistilBERT & AG News & \textit{BAE} & 55.8 & 44.0 & \textbf{86.4} & \textbf{94.5} & 85.8 & 93.2 & 85.3 & 92.8 \\
             DistilBERT & RTMR & \textit{BAE} & 29.4 & 18.5 & 71.0 & 75.2 & \textbf{79.1}  & \textbf{80.8}  & 65.5 & 60.8 \\
             \hline
             \multicolumn{3}{ c |}{Overall average} & 77.1 & 68.8 & 86.4 & 89.4 & 92.4 & 95.2 & - & -
        \end{tabular}
        \caption{$\mathrm{FGWS}$ and $\mathrm{WDR}$ trained on configuration (DistilBERT, AG News, \textit{PWWS}).}
        \label{tab:comparision_AG}
    \end{subtable}

     \caption{Comparison between $\mathrm{VoteTRANS}$ and two state-of-the-art detectors, $\mathrm{FGWS}$ and $\mathrm{WDR}$. 
     While both $\mathrm{FGWS}$ and $\mathrm{WDR}$ need to be trained on a specific configuration (model, data, \textit{attack}), $\mathrm{VoteTRANS}$ directly detects adversarial text without training. 
     $\mathrm{VoteTRANS_{same}}$ uses an auxiliary attack that is the same as the attack used to generate adversarial texts.
     If the adversarial text is not generated by \textit{PWWS}, $\mathrm{VoteTRANS_{diff}}$ uses \textit{PWWS} as the auxiliary attack.
     }
     \label{tab:comparison}
\end{table*}

$\mathrm{VoteTRANS}$ uses an auxiliary attack to detect adversarial text without training.
The auxiliary attack may be the same as the attack used to generate the adversarial text, namely, $\mathrm{VoteTRANS_{same}}$.
We also demonstrate the capability of $\mathrm{VoteTRANS}$ using a fixed auxiliary attack (\textit{PWWS}) to detect adversarial text generated from other attacks, namely, $\mathrm{VoteTRANS_{diff}}$.
Both $\mathrm{VoteTRANS_{same}}$ and $\mathrm{VoteTRANS_{diff}}$ uses RoBERTa as a support for all experiments\footnote{Since \mbox{TextAttack} does not support RoBERTa for YELP, we used ALBERT as the support instead.}.
Other auxiliary attacks and supports will be discussed later.

\begin{table*}[!ht]
    \centering
    \begin{tabular}{l | l|c c}
         \textbf{Scenario} & \textbf{Method}& \textbf{F1} & \textbf{Recall}\\
         \hline
        \multirow{15}{*}{Unknown Attack} & $\mathrm{VoteTRANS_{diff}}$ (\textit{BAE}) without support & 94.4 & 93.6\\
                                        & $\mathrm{VoteTRANS_{diff}}$ (\textit{DeepWordBug}) without support & 94.2 & 97.2\\
                                        & $\mathrm{VoteTRANS_{diff}}$ (\textit{BAE}) with RoBERTa as support & \textbf{96.7} & 99.6\\
                                        & $\mathrm{VoteTRANS_{diff}}$ (\textit{DeepWordBug}) with RoBERTa as support & 95.2 & \textbf{100.0}\\
                                        & $\mathrm{VoteTRANS_{diff}}$ (\textit{Checklist}) with RoBERTa as support & 83.9 & 74.0\\
                                        &  $\mathrm{VoteTRANS_{diff}}$ (\textit{Input-Reduction}) with RoBERTa as support & 92.8 & \textbf{100.0}\\
                                        &  $\mathrm{VoteTRANS_{diff}}$ (\textit{A2T}) with RoBERTa as support & 95.7 & 96.8\\
                                        &  $\mathrm{VoteTRANS_{diff}}$ (\textit{IGA}) with RoBERTa as support & 95.7 & 97.2\\
                                        &  $\mathrm{VoteTRANS_{diff}}$ (\textit{Pruthi}) with RoBERTa as support & 95.8 & 99.6\\
                                        &  $\mathrm{VoteTRANS_{diff}}$ (\textit{Alzantot}) with RoBERTa as support & 95.9 & 97.6\\
                                        &  $\mathrm{VoteTRANS_{diff}}$ (\textit{PSO}) with RoBERTa as support & 96.1 & 97.6\\
                                        &  $\mathrm{VoteTRANS_{diff}}$ (\textit{Faster-Alzantot}) with RoBERTa as support & 96.4 & 97.6\\
                                        &  $\mathrm{VoteTRANS_{diff}}$ (\textit{TextBugger}) with RoBERTa as support & 96.5 & 98.8\\
                                        &  $\mathrm{VoteTRANS_{diff}}$ (\textit{TextFooler}) with RoBERTa as support & 96.5 & 98.0\\
                                        &  $\mathrm{VoteTRANS_{diff}}$ (\textit{Kuleshov}) with RoBERTa as support & 96.6 & 97.6\\
        \hline
        \multirow{6}{*}{Known Attack}   & $\mathrm{FGWS}$ & 90.6 & 87.6\\
                                        & $\mathrm{WDR}$ & 91.1 & 91.2\\
                                        & $\mathrm{VoteTRANS_{same}}$ (\textit{PWWS}) without support & 95.2 & 94.8\\
                                        & $\mathrm{VoteTRANS_{same}}$ (\textit{PWWS}) with LSTM as support & 95.9 & 98.8\\
                                        & $\mathrm{VoteTRANS_{same}}$ (\textit{PWWS}) with RoBERTa as support & 96.5 & \textbf{99.2}\\
                                        & $\mathrm{VoteTRANS_{same}}$ (\textit{PWWS}) with LSTM+RoBERTa as supports & \textbf{97.4} & 98.4\\

    \end{tabular}
    \caption{Detecting adversarial text generated by \textit{PWWS} targeting CNN on AG News.}
    \label{tab:ablation_studies}
\end{table*}

In general, $\mathrm{WDR}$ is better than $\mathrm{FGWS}$ when they are trained on (DistilBERT, IMDB, \textit{PWWS}) as shown in Table~\ref{tab:comparision_IMDB}. 
While $\mathrm{FGWS}$ achieves an F1 score of 77.5 and recall of 68.4 on average, $\mathrm{WDR}$ exhibits improved F1 score and recall metrics of 85.4 and 91.7, respectively.
All F1 scores of $\mathrm{VoteTRANS_{same}}$ outperform those of $\mathrm{WDR}$.
$\mathrm{VoteTRANS_{same}}$ also exhibits a recall improved by 4.0 points from 91.7 of $\mathrm{WDR}$.
$\mathrm{VoteTRANS_{diff}}$ is competitive with $\mathrm{VoteTRANS_{same}}$ for medium (AG News) and long text (IMDB), while the performance of $\mathrm{VoteTRANS_{diff}}$ on short text (RTMR) is degraded, similar to other detectors.

\begin{table*}[!ht]
    \setlength\tabcolsep{1.0pt} 
    \centering
    \begin{tabular}{l|c  c c}
        \textbf{Category} & \textbf{RTMR(Adv/Org)} & \textbf{AG News(Adv/Org)} & \textbf{IMDB(Adv/Org)} \\
        \hline
         \textit{PWWS} attack time & 0.77 & 2.84 & 26.36 \\
         $\mathrm{FGWS}$/$\mathrm{WDR}$ & 0.04 & 0.08 & 1.01 \\
         $\mathrm{VoteTRANS_{same}}$ without support & 0.03(0.02/0.04) & 0.08(0.03/0.13) & 2.00(0.67/3.33) \\
        $\mathrm{VoteTRANS_{same}}$ with RoBERTa as support & 0.69(0.28/1.09) & 1.42(0.15/2.69) & 8.94(0.37/17.52) \\ 
    \end{tabular}
    \caption{Run time for attacking original text by \textit{PWWS} and detecting adversarial text generated by \textit{PWWS} targeting the CNN model.}
    \label{tab:running_time}
\end{table*}

In detail, we cluster the experimental results into three groups based on their performances.
The first group with high performances includes configurations from similar attacks (\textit{PWWS}, \textit{TextFooler}, and \textit{IGA}) on long (IMDB and YELP) and medium text (AG News).
The second group includes configurations from \textit{PWWS} and \textit{IGA} on short text (RTMR).
The last group includes the remaining configurations related to \textit{BAE}.
While \textit{BAE} uses flexible synonyms based on word context, the other attacks use fixed synonyms for a certain word.

All of the detectors work well for the lengthy text of the first group, especially $\mathrm{VoteTRANS_{same}}$, with both scores being between 94.8 and 99.6.
However, less information can be extracted from the short text in the second group.
While $\mathrm{FGWS}$ is competitive with $\mathrm{WDR}$ in this group, $\mathrm{VoteTRANS_{same}}$ performs better, especially in the F1 score.
In the last group, \textit{BAE} remarkably affects the detectors, especially with $\mathrm{FGWS}$ in medium and short text.
$\mathrm{FGWS}$ is defeated, with scores less than 50.0 (random guess).
$\mathrm{VoteTRANS_{same}}$ still maintains its high performance for long text and is competitive with $\mathrm{WDR}$ for medium and short text.

Table~\ref{tab:comparision_AG} shows experiments where $\mathrm{FGWS}$ and $\mathrm{WDR}$ are trained on (DistilBERT, AG News, \textit{PWWS}). 
We train $\mathrm{WDR}$ with other word-based attacks including \textit{TextFooler}, \textit{IGA}, 
and \textit{BAE} and reach similar results as shown in Appendix~\ref{sec:appendix:word_based_attacks_training}.
While $\mathrm{FGWS}$ and $\mathrm{WDR}$ obtain scores less than 90.0 on average, $\mathrm{VoteTRANS_{same}}$ retains its performance, with 92.4 F1 and 95.2 recall.
This demonstrates the resilience of $\mathrm{VoteTRANS_{same}}$ across various models, datasets, and attacks.

\subsection{Ablation Studies}

We studied variants of $\mathrm{VoteTRANS}$ and compared them with $\mathrm{FGWS}$ and $\mathrm{WDR}$.
These detectors identified adversarial texts generated by \textit{PWWS} targeting the CNN on AG News (Table~\ref{tab:ablation_studies}).
$\mathrm{VoteTRANS}$ is presented in two scenarios: an unknown attack ($\mathrm{VoteTRANS_{diff}}$) and a known attack ($\mathrm{VoteTRANS_{same}}$).

For an unknown attack, we use a word-based \textit{BAE} and character-based \textit{DeepWordBug} as the auxiliary attacks for $\mathrm{VoteTRANS_{diff}}$ without support.  
$\mathrm{VoteTRANS_{diff}}$ achieves high performances with both auxiliaries.
Other auxiliaries for $\mathrm{VoteTRANS_{diff}}$ without support are mentioned in Appendix~\ref{sec:appendix:other_diff_without_support}.
The use of the RoBERTa model as support boosts the overall performance.
Other attacks from the \mbox{TextAttack} were conducted and are listed in increasing order of F1 scores.
Among these attacks, \textit{BERT-Attack} and CLARE are ignored because both use the same masked language model used in \textit{BAE}, and the three attacks reached similar performances.
\textit{HotFlip} is not supported for CNN.
The results show that $\mathrm{VoteTRANS_{diff}}$ can use any attack as the auxiliary, with all scores being greater than or equal to 92.8, except for those of \textit{Checklist}.
\textit{Checklist} generates independent adversarial text with any model and causes low performance, as mentioned in the owner paper~\citep{ribeiro2020beyond}.
The results from various auxiliaries demonstrate that $\mathrm{VoteTRANS_{diff}}$ can detect adversarial text without attack information.

For a known attack, $\mathrm{VoteTRANS_{same}}$ without support outperforms $\mathrm{FGWS}$ and $\mathrm{WDR}$.
$\mathrm{VoteTRANS_{same}}$ is improved by using random support such as LSTM or RoBERTa.
A stronger model (RoBERTa) helps $\mathrm{VoteTRANS_{same}}$ more than LSTM.
Both can also be used together to support $\mathrm{VoteTRANS_{same}}$ and improve the F1 score, but the adversarial recall is slightly affected.
Other available supports from \mbox{TextAttack} for AG News are mentioned in Appendix~\ref{sec:appendix:other_supports}.

While character-based attacks are still a challenge for both $\mathrm{WDR}$ and $\mathrm{FGWS}$ as mentioned in their papers, $\mathrm{VoteTRANS}$ still detects such adversarial text upto 97.6\% F1 and 99.6\% recall as shown in Appendix~\ref{sec:appendix:char_attacks}. 
It indicates the flexible of $\mathrm{VoteTRANS}$ with various attack levels.

\subsection{Run Time}
\label{sec:run_time}

Since $\mathrm{VoteTRANS}$ uses an auxiliary attack to detect adversarial text, we compared the run time of $\mathrm{VoteTRANS_{same}}$ with that of the corresponding attack.
Table~\ref{tab:running_time} shows a comparison of adversarial text generated by \textit{PWWS} targeting the CNN model on short text (RTMR), medium text (AG News), and long text (IMDB); $\mathrm{VoteTRANS_{diff}}$, other attacks, and models reached similar ratios.
We also compared the detection times obtained from $\mathrm{WDR}$/$\mathrm{FGWS}$, which both use the target model to predict the text $n$ times, where $n$ is the number of words in an input text.
$\mathrm{VoteTRANS_{same}}$ is reported without support and with RoBERTa support.
We also separately appended the detection time for adversarial and original text for $\mathrm{VoteTRANS_{same}}$, while other detectors ran for the same time for both.

The run times of both the attack and detectors are affected by the text length.
$\mathrm{FGWS}$ and $\mathrm{WDR}$ need less than 2 seconds to detect text.
$\mathrm{VoteTRANS_{same}}$ processes adversarial text much faster than original text because most of the adversarial text is identified early with lines 20-22 of Algorithm~\ref{alg:proposal}.
Thanks to line 12 of Algorithm~\ref{alg:proposal} for filtering many transformed texts, $\mathrm{VoteTRANS_{same}}$ without support run similar to $\mathrm{FGWS}$ and $\mathrm{WDR}$ for short and medium text.
For long text, $\mathrm{VoteTRANS_{same}}$ needs 0.67 seconds for adversarial text.
$\mathrm{VoteTRANS_{same}}$ with RoBERTa as support even completes processing the adversarial text from IMDB with only 0.37 seconds.
This demonstrates that RoBERTa accelerates adversarial text processing.

$\mathrm{VoteTRANS_{same}}$ runs faster by decreasing the word ratio $\alpha$ in Algorithm~\ref{alg:proposal}. 
$\alpha$ determines the number of words that are processed.
While $\mathrm{VoteTRANS_{same}}$ with RoBERTa as support processes RTMR text in a reasonable time, we evaluate the change in $\alpha$ for processing AG News and IMDB text, as shown in Figure~\ref{fig:threshold_AG} and Figure~\ref{fig:threshold_IMDB}, respectively.
For AG News, although the detection time is mostly steady at 0.25 seconds, with $\alpha$ between 12\% and 19\%, the F1/recall ratio remarkably increases from 93.9/91.6 to 95.4/95.2.
The run time is worsened to 1.42 seconds with the largest $\alpha$ of 100\%, but the F1 scores are only slightly increased.

\begin{figure}[]
    \begin{subfigure}[]{0.5\textwidth}
    \centering
    \includegraphics[]{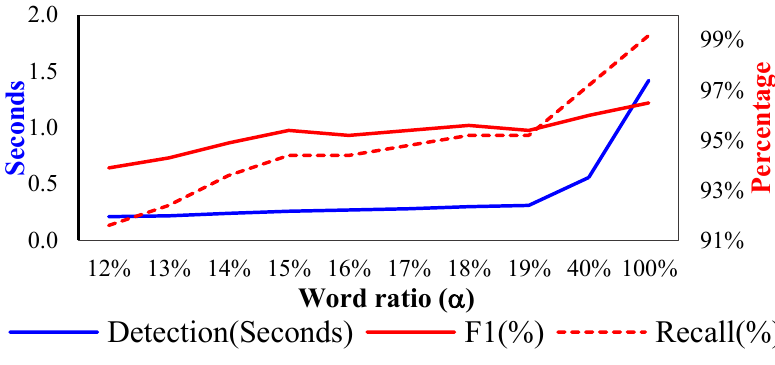}
    \caption{AG News.}
    \label{fig:threshold_AG}
    \end{subfigure}
    \begin{subfigure}[b]{0.5\textwidth}
    \centering
    \includegraphics[]{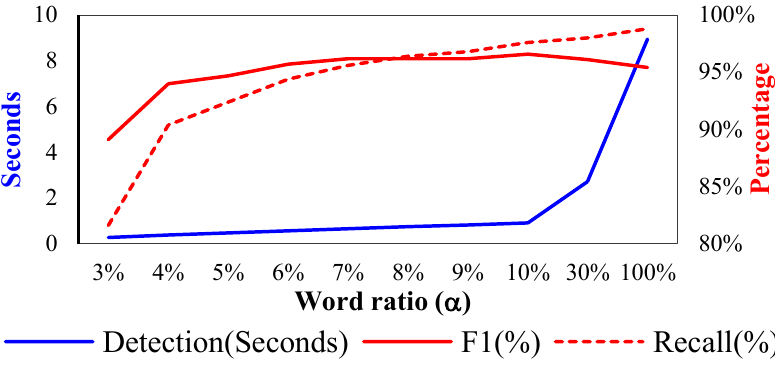}
    \caption{IMDB.}
    \label{fig:threshold_IMDB}
    \end{subfigure}

    \centering
    \caption{Correlation between the detection time and performance of $\mathrm{VoteTRANS_{same}}$ with RoBERTa as support to detect adversarial text generated by \textit{PWWS} targeting the CNN model when changing the word ratio $\alpha$. The time is averaged for all original and adversarial detection.}
    \label{fig:my_label}
\end{figure}

For IMDB text, $\mathrm{VoteTRANS_{same}}$ achieves a high performance, even with a small $\alpha$.
When $\alpha$ is increased from 3\% to 10\%, the run time/F1/recall increases from 0.28/89.1/81.6 to 0.91/96.6/97.6.
The recall is improved up to 98.8 with a maximum $\alpha$ of 100\%, but the corresponding F1 score drops slightly to 95.4.
The F1 score is affected by some of the original text misclassified as adversarial text.

In particular, when processing with 12\% and 5\% of the medium text (AG News) and long text (IMDB), respectively, $VoteTRANS_{same}$ with RoBERTa as support takes approximately 0.21 seconds and 0.47 seconds, competitive with 0.08 seconds and 1.01 seconds produced from FGWS/WDR, while keeping F1/recall scores (93.9/91.6 for AG News and 94.7/92.4 for IMDB), higher than FGWS (90.6/87.6 and 86.3/79.6) and WDR (91.1/91.2 and 84.3/90.0).
With these ratios,
$VoteTRANS_{same}$ with RoBERTa only needs 32.0 and 61.1 for AG News and IMDB, respectively (see Appendix~\ref{sec:appendix:number_predictions} for other ratios).

\subsection{Discussion}

\textbf{Detection with high confidence text:}
$\mathrm{VoteTRANS}$ still keeps 79.2\% of F1 score on detecting adversarial text
from AG News with its confidence greater than or equal to 90\%.
In contrast, $\mathrm{WDR}$ drops the score to 25.0\% (see other confidences and IMDB text in Appendix~\ref{sec:appendix:high_confidence_adv}).

\textbf{Detection with only hard labels:}
$\mathrm{VoteTRANS}$ only uses soft labels of predictions from a target model via an auxiliary attack to calculate importance scores (line 3 of Algorithm~\ref{alg:proposal}).
However, these scores are only used to accelerate detection by selecting the top words in line 6.
Without these scores, $\mathrm{VoteTRANS}$ achieves an identical performance by processing all words.
Therefore, $\mathrm{VoteTRANS}$ is compatible with any target model that only provides hard labels.

\textbf{Parallel processing:}
An adversarial attack needs to perturb individual words of input text in sequence to optimize the perturbed text in each step until a target model is fooled.
On the other hand, $\mathrm{VoteTRANS}$ can create independent transformation sets for individual words and process them in parallel.
$\mathrm{VoteTRANS}$ can accelerate the process with parallel or distributed computing.

\begin{table}[t]
    \centering
    \begin{tabular}{l|r r}
         \textbf{Configuration}& \textbf{AG News} & \textbf{IMDB}\\
         \hline
         Without detector & 88.8 & 100.0\\
         With $\mathrm{WDR}$ & 5.2 & 16.4\\
         With $\mathrm{VoteTRANS}$ & \textbf{0.4} & \textbf{4.3} \\

    \end{tabular}
    \caption{Success rate under an adaptive attack.}
    \label{tab:adaptive_attack}
\end{table}

\textbf{Adaptive attack:}
An attacker may be aware 
of the existence of a detector and fool both a target and the detector. 
We evaluate \textit{PWWS} targeting CNN models on AG News and IMDB as shown in Table~\ref{tab:adaptive_attack}.
Although \textit{PWWS} strongly attacks the CNN models with more than 88\% of success rate, it hardly bypasses detectors, especially $\mathrm{VoteTRANS}$.

\section{Conclusion}

We propose $\mathrm{VoteTRANS}$, a method for detecting adversarial text without training by voting on hard labels of text after transformation.
$\mathrm{VoteTRANS}$ outperforms state-of-the-art detectors under various attacks, models, and datasets.
Moreover, $\mathrm{VoteTRANS}$ is flexible at detecting a restricted scenario when an attack is unknown.
$\mathrm{VoteTRANS}$ also straightforwardly detects adversarial text from a new attack without modifying the architecture.

\section{Limitations}

\paragraph{Auxiliary attack and supports:}
$\mathrm{VoteTRANS}$ without support works well with an auxiliary attack which is the same with the target attack.
In contrast,  $\mathrm{VoteTRANS}$ with support achieves stable results with any auxiliary attack but it runs slower.
\paragraph{Short text and susceptible text:} 
A short text is more difficult to detect than a long text.
Susceptible text may bypass $\mathrm{VoteTRANS}$ as mentioned in Appendix~\ref{sect_error_analysis}.
However, the short text and susceptible text are often unnatural and unclear meaning, respectively, so they are easily recognized by humans.
Therefore, we recommend that humans recheck suspicious text with an abnormal ratio in the voting process of $\mathrm{VoteTRANS}$ (line 19 of Algorithm~\ref{alg:proposal}).

\paragraph{Beyond word-based attacks:} 
We detect adversarial text up to word-based attacks, which change a few characters or words and are often imperceptible to humans.
Other attacks remarkably affect the naturalness with a large change
such as sentence-based attacks as in \citet{iyyer2018adversarial}.

\paragraph{Beyond text classification:} 
We evaluate $\mathrm{VoteTRANS}$ on adversarial attacks targeting text classification.
In contrast, the other tasks do not well-define a standard for generating adversarial text.
For example, attacks targeting sequence models need to determine a threshold for $BLEU$ score, which is aimed to minimize, but whether the score is sufficient for an adversarial text is still in question.

\section*{Acknowledgments}

This work was partially supported by JST CREST Grants JPMJCR20D3, Japan. 

\bibliography{votetrans}

\begin{thebibliography}{27}
\expandafter\ifx\csname natexlab\endcsname\relax\def\natexlab#1{#1}\fi

\bibitem[{Alzantot et~al.(2018)Alzantot, Sharma, Elgohary, Ho, Srivastava, and
  Chang}]{alzantot2018generating}
Moustafa Alzantot, Yash Sharma, Ahmed Elgohary, Bo-Jhang Ho, Mani Srivastava,
  and Kai-Wei Chang. 2018.
\newblock \href {https://aclanthology.org/D18-1316.pdf} {Generating natural
  language adversarial examples}.
\newblock In \emph{Proceedings of the Conference on Empirical Methods in
  Natural Language Processing (EMNLP)}, pages 2890--2896.

\bibitem[{Biju et~al.(2022)Biju, Sriram, Kumar, and Khapra}]{biju2022input}
Emil Biju, Anirudh Sriram, Pratyush Kumar, and Mitesh~M Khapra. 2022.
\newblock \href {https://aclanthology.org/2022.findings-acl.4/} {Input-specific
  attention subnetworks for adversarial detection}.
\newblock In \emph{Findings of the Association for Computational Linguistics
  (ACL)}, pages 31--44.

\bibitem[{Ebrahimi et~al.(2018)Ebrahimi, Rao, Lowd, and
  Dou}]{ebrahimi2018hotflip}
Javid Ebrahimi, Anyi Rao, Daniel Lowd, and Dejing Dou. 2018.
\newblock \href {https://aclanthology.org/P18-2006.pdf} {Hotflip: White-box
  adversarial examples for text classification}.
\newblock In \emph{Proceedings of the 56th Annual Meeting of the Association
  for Computational Linguistics (ACL)}, pages 31--36.

\bibitem[{Feng et~al.(2018)Feng, Wallace, Grissom~II, Rodriguez, Iyyer, and
  Boyd-Graber}]{feng2018pathologies}
Shi Feng, Eric Wallace, Alvin Grissom~II, Pedro Rodriguez, Mohit Iyyer, and
  Jordan Boyd-Graber. 2018.
\newblock \href {https://aclanthology.org/D18-1407} {Pathologies of neural
  models make interpretation difficult}.
\newblock In \emph{Proceedings of the Conference on Empirical Methods in
  Natural Language Processing (EMNLP)}, pages 3719--3728.

\bibitem[{Gao et~al.(2018)Gao, Lanchantin, Soffa, and Qi}]{gao2018black}
Ji~Gao, Jack Lanchantin, Mary~Lou Soffa, and Yanjun Qi. 2018.
\newblock \href {https://ieeexplore.ieee.org/document/8424632} {Black-box
  generation of adversarial text sequences to evade deep learning classifiers}.
\newblock In \emph{Proceedings of the IEEE Security and Privacy Workshops
  (SPW)}, pages 50--56.

\bibitem[{Garg and Ramakrishnan(2020)}]{garg2020bae}
Siddhant Garg and Goutham Ramakrishnan. 2020.
\newblock \href {https://aclanthology.org/2020.emnlp-main.498} {Bae: Bert-based
  adversarial examples for text classification}.
\newblock In \emph{Proceedings of the Conference on Empirical Methods in
  Natural Language Processing (EMNLP)}, pages 6174--6181.

\bibitem[{Iyyer et~al.(2018)Iyyer, Wieting, Gimpel, and
  Zettlemoyer}]{iyyer2018adversarial}
Mohit Iyyer, John Wieting, Kevin Gimpel, and Luke Zettlemoyer. 2018.
\newblock \href {https://aclanthology.org/N18-1170.pdf} {Adversarial example
  generation with syntactically controlled paraphrase networks}.
\newblock In \emph{Proceedings of the 16th Conference of the North American
  Chapter of the Association for Computational Linguistics (NAACL)}, pages
  1875--1885.

\bibitem[{Jia et~al.(2019)Jia, Raghunathan, G{\"o}ksel, and
  Liang}]{jia2019certified}
Robin Jia, Aditi Raghunathan, Kerem G{\"o}ksel, and Percy Liang. 2019.
\newblock \href {https://aclanthology.org/D19-1423.pdf} {Certified robustness
  to adversarial word substitutions}.
\newblock In \emph{Proceedings of the Conference on Empirical Methods in
  Natural Language Processing (EMNLP)}, pages 4120--4133.

\bibitem[{Jin et~al.(2020)Jin, Jin, Tianyi~Zhou, and Szolovits}]{jin2020bert}
Di~Jin, Zhijing Jin, Joey Tianyi~Zhou, and Peter Szolovits. 2020.
\newblock \href {https://aaai.org/ojs/index.php/AAAI/article/view/6311} {Is
  bert really robust? a strong baseline for natural language attack on text
  classification and entailment}.
\newblock In \emph{Proceedings of the 34th Conference on Artificial
  Intelligence (AAAI)}, pages 8018--8025.

\bibitem[{Kuleshov et~al.(2018)Kuleshov, Thakoor, Lau, and
  Ermon}]{kuleshov2018adversarial}
Volodymyr Kuleshov, Shantanu Thakoor, Tingfung Lau, and Stefano Ermon. 2018.
\newblock \href {https://openreview.net/forum?id=r1QZ3zbAZ} {Adversarial
  examples for natural language classification problems}.
\newblock In \emph{OpenReview}.

\bibitem[{Li et~al.(2021)Li, Zhang, Peng, Chen, Brockett, Sun, and
  Dolan}]{li2021contextualized}
Dianqi Li, Yizhe Zhang, Hao Peng, Liqun Chen, Chris Brockett, Ming-Ting Sun,
  and William~B Dolan. 2021.
\newblock \href {https://aclanthology.org/2021.naacl-main.400} {Contextualized
  perturbation for textual adversarial attack}.
\newblock In \emph{Proceedings of the Conference of the North American Chapter
  of the Association for Computational Linguistics (NAACL)}, pages 5053--5069.

\bibitem[{Li et~al.(2019)Li, Ji, Du, Li, and Wang}]{li2019textbugger}
J~Li, S~Ji, T~Du, B~Li, and T~Wang. 2019.
\newblock \href
  {https://www.ndss-symposium.org/wp-content/uploads/2019/02/ndss2019_03A-5_Li_paper.pdf}
  {Textbugger: Generating adversarial text against real-world applications}.
\newblock In \emph{Proceedings of the 26th Annual Network and Distributed
  System Security Symposium (NDSS)}.

\bibitem[{Li et~al.(2020)Li, Ma, Guo, Xue, and Qiu}]{li2020bert}
Linyang Li, Ruotian Ma, Qipeng Guo, Xiangyang Xue, and Xipeng Qiu. 2020.
\newblock \href {https://aclanthology.org/2020.emnlp-main.500.pdf}
  {Bert-attack: Adversarial attack against bert using bert}.
\newblock In \emph{Proceedings of the Conference on Empirical Methods in
  Natural Language Processing (EMNLP)}, pages 6193--6202.

\bibitem[{Morris et~al.(2020)Morris, Lifland, Yoo, Grigsby, Jin, and
  Qi}]{morris2020textattack}
John Morris, Eli Lifland, Jin~Yong Yoo, Jake Grigsby, Di~Jin, and Yanjun Qi.
  2020.
\newblock \href {https://aclanthology.org/2020.emnlp-demos.16.pdf} {Textattack:
  A framework for adversarial attacks, data augmentation, and adversarial
  training in nlp}.
\newblock In \emph{Proceedings of the Conference on Empirical Methods in
  Natural Language Processing: System Demonstrations (EMNLP)}, pages 119--126.

\bibitem[{Mosca et~al.(2022)Mosca, Agarwal, Rando-Ramirez, and
  Groh}]{mosca2022suspicious}
Edoardo Mosca, Shreyash Agarwal, Javier Rando-Ramirez, and Georg Groh. 2022.
\newblock \href {https://aclanthology.org/2022.acl-long.538/} {" that is a
  suspicious reaction!": Interpreting logits variation to detect nlp
  adversarial attacks}.
\newblock In \emph{Proceedings of the 60th Annual Meeting of the Association
  for Computational Linguistics (ACL)}, pages 7806--7816.

\bibitem[{Mozes et~al.(2021)Mozes, Stenetorp, Kleinberg, and
  Griffin}]{mozes2021frequency}
Maximilian Mozes, Pontus Stenetorp, Bennett Kleinberg, and Lewis Griffin. 2021.
\newblock \href {https://aclanthology.org/2021.eacl-main.13.pdf}
  {Frequency-guided word substitutions for detecting textual adversarial
  examples}.
\newblock In \emph{Proceedings of the 16th Conference of the European Chapter
  of the Association for Computational Linguistics (EACL)}, pages 171--186.

\bibitem[{Pruthi et~al.(2019)Pruthi, Dhingra, and Lipton}]{pruthi2019combating}
Danish Pruthi, Bhuwan Dhingra, and Zachary~C Lipton. 2019.
\newblock \href {http://aclanthology.lst.uni-saarland.de/P19-1561.pdf}
  {Combating adversarial misspellings with robust word recognition}.
\newblock In \emph{Proceedings of the 57th Annual Meeting of the Association
  for Computational Linguistics (ACL)}, pages 5582--5591.

\bibitem[{Raina and Gales(2022)}]{raina2022residue}
Vyas Raina and Mark Gales. 2022.
\newblock \href {https://aclanthology.org/2022.naacl-main.281.pdf}
  {Residue-based natural language adversarial attack detection}.
\newblock In \emph{Proceedings of the Conference of the North American Chapter
  of the Association for Computational Linguistics (NAACL)}, pages 3836--3848.

\bibitem[{Ren et~al.(2019)Ren, Deng, He, and Che}]{ren2019generating}
Shuhuai Ren, Yihe Deng, Kun He, and Wanxiang Che. 2019.
\newblock \href {https://aclanthology.org/P19-1103} {Generating natural
  language adversarial examples through probability weighted word saliency}.
\newblock In \emph{Proceedings of the 57th Annual Meeting of the Association
  for Computational Linguistics (ACL)}, pages 1085--1097.

\bibitem[{Ribeiro et~al.(2020)Ribeiro, Wu, Guestrin, and
  Singh}]{ribeiro2020beyond}
Marco~Tulio Ribeiro, Tongshuang Wu, Carlos Guestrin, and Sameer Singh. 2020.
\newblock \href {https://aclanthology.org/2020.acl-main.442} {Beyond accuracy:
  Behavioral testing of nlp models with checklist}.
\newblock In \emph{Proceedings of the 58th Annual Meeting of the Association
  for Computational Linguistics (ACL)}, pages 4902--4912.

\bibitem[{Wang et~al.(2022{\natexlab{a}})Wang, Bao, Zhang, and
  Zhao}]{wang2022distinguishing}
Jiayi Wang, Rongzhou Bao, Zhuosheng Zhang, and Hai Zhao. 2022{\natexlab{a}}.
\newblock \href {https://aclanthology.org/2022.findings-acl.73} {Distinguishing
  non-natural from natural adversarial samples for more robust pre-trained
  language model}.
\newblock In \emph{Findings of the Association for Computational Linguistics
  (ACL)}, pages 905--915.

\bibitem[{Wang et~al.(2021)Wang, Jin, Yang, and He}]{wang2021natural}
Xiaosen Wang, Hao Jin, Yichen Yang, and Kun He. 2021.
\newblock \href {https://www.auai.org/uai2021/pdf/uai2021.315.pdf} {Natural
  language adversarial defense through synonym encoding}.
\newblock In \emph{Proceedings of the 37th Conference on Uncertainty in
  Artificial Intelligence (UAI)}, pages 823--833.

\bibitem[{Wang et~al.(2022{\natexlab{b}})Wang, Xiong, and
  He}]{wang2021randomized}
Xiaosen Wang, Yifeng Xiong, and Kun He. 2022{\natexlab{b}}.
\newblock \href {https://arxiv.org/abs/2109.05698} {Randomized substitution and
  vote for textual adversarial example detection}.
\newblock In \emph{Proceedings of the 38th Conference on Uncertainty in
  Artificial Intelligence (UAI)}.

\bibitem[{Yoo and Qi(2021)}]{yoo2021towards}
Jin~Yong Yoo and Yanjun Qi. 2021.
\newblock \href {https://aclanthology.org/2021.findings-emnlp.81.pdf} {Towards
  improving adversarial training of {NLP} models}.
\newblock In \emph{Findings of the Association for Computational Linguistics
  (EMNLP)}, pages 945--956.

\bibitem[{Yoo et~al.(2022)Yoo, Kim, Jang, and Kwak}]{yoo2022detection}
KiYoon Yoo, Jangho Kim, Jiho Jang, and Nojun Kwak. 2022.
\newblock \href {https://aclanthology.org/2022.findings-acl.289} {Detection of
  word adversarial examples in text classification: Benchmark and baseline via
  robust density estimation}.
\newblock In \emph{Findings of the 60th Annual Meeting of the Association for
  Computational Linguistics (ACL)}, pages 3656--3672.

\bibitem[{Zang et~al.(2020)Zang, Qi, Yang, Liu, Zhang, Liu, and
  Sun}]{zang2020word}
Yuan Zang, Fanchao Qi, Chenghao Yang, Zhiyuan Liu, Meng Zhang, Qun Liu, and
  Maosong Sun. 2020.
\newblock \href {https://aclanthology.org/2020.acl-main.540.pdf} {Word-level
  textual adversarial attacking as combinatorial optimization}.
\newblock In \emph{Proceedings of the 58th Annual Meeting of the Association
  for Computational Linguistics (ACL)}, pages 6066--6080.

\bibitem[{Zhou et~al.(2019)Zhou, Jiang, Chang, and Wang}]{zhou2019learning}
Yichao Zhou, Jyun-Yu Jiang, Kai-Wei Chang, and Wei Wang. 2019.
\newblock \href {https://aclanthology.org/D19-1496.pdf} {Learning to
  discriminate perturbations for blocking adversarial attacks in text
  classification}.
\newblock In \emph{Proceedings of the Conference on Empirical Methods in
  Natural Language Processing (EMNLP)}, pages 4906--4915.

\end{thebibliography}
\bibliographystyle{acl_natbib}

\appendix

\section{Prediction Change under One-Word Transformation}
\label{section:appendix:max_rate}
\begin{figure}[]
    \begin{subfigure}[]{0.5\textwidth}
    \centering
    \includegraphics[]{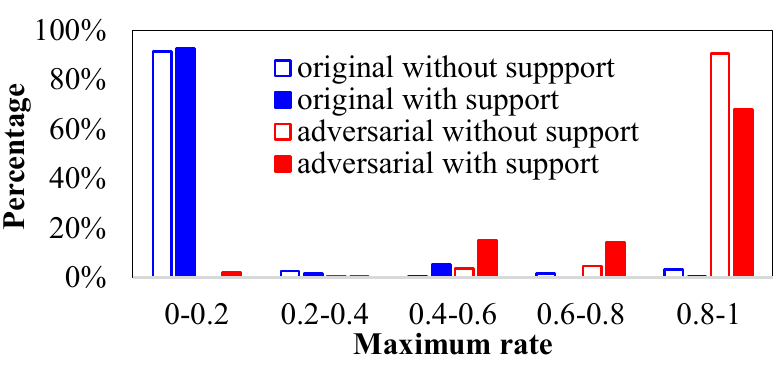}
    \caption{AG News.}
    \label{fig:threshold_AG}
    \end{subfigure}
    \begin{subfigure}[b]{0.5\textwidth}
    \centering
    \includegraphics[]{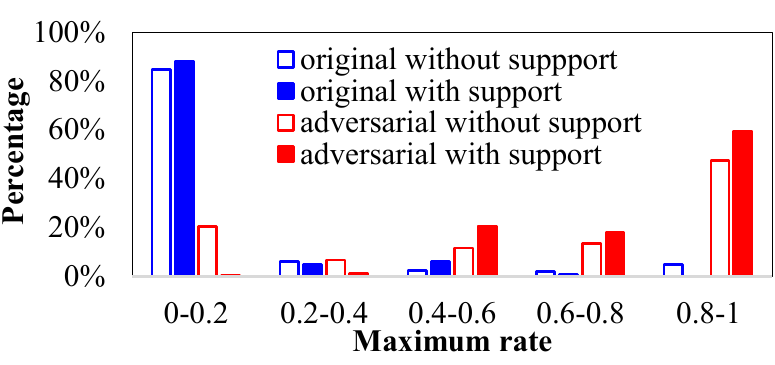}
    \caption{IMDB.}
    \label{fig:threshold_IMDB}
    \end{subfigure}

    \centering
    \caption{Prediction change under one-word transformation.}
    \label{fig:max_rate}
\end{figure}

The example in Figure~\ref{fig:example} shows that the prediction of an adversarial text is more susceptible than that of an original text.
We verify this observation on other AG News texts and IMDB movie reviews.
In particular, we inspect the whole testing set containing 500 balanced samples of original and adversarial text generated by \textit{PWWS} targeting CNN.
For each sample, we transform a word by synonyms and measure the rate of prediction change from the target CNN.
The maximum rate among words is represented for the sample and is plotted in histogram graphs (Figure~\ref{fig:max_rate}).
The maximum rate of an original text is often lower than that of an adversarial text in both AG News and IMDB.

\section{Word Importance Score}
\label{sec:appendix:word_importance}

\begin{figure}[!ht]
    \begin{subfigure}[b]{0.5\textwidth}
    \centering
    \includegraphics[]{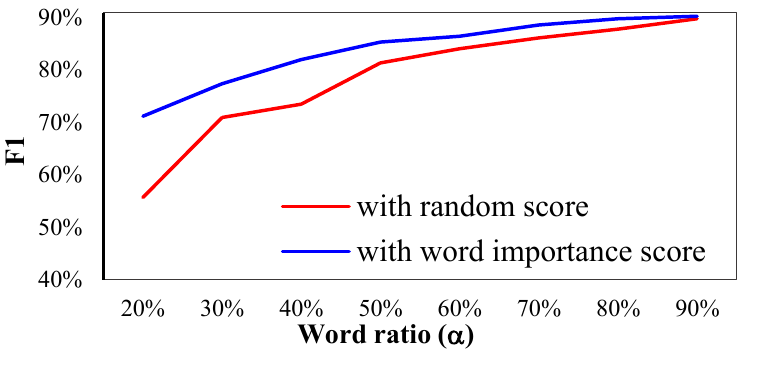}
    \caption{F1.}
    \label{fig:word_importance_F1}
    \end{subfigure}
    \begin{subfigure}[b]{0.5\textwidth}
    \centering
    \includegraphics[]{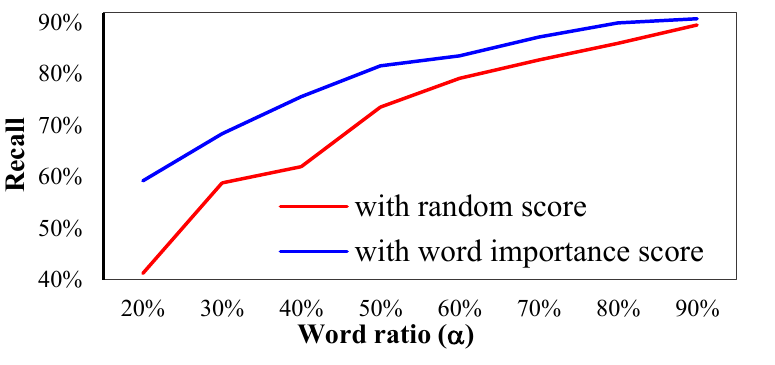}
    \caption{Recall.}
    \label{fig:word_importance_recall}
    \end{subfigure}

    \centering
    \caption{The impact of word importance score in $\mathrm{VoteTRANS}$ via changing word ratio $\alpha$.}
    \label{fig:word_importace}
\end{figure}

We change word ratio $\alpha$ with 10\% step to show the impact of word importance score in $\mathrm{VoteTRANS}$ (line 2-6 in Algorithm~\ref{alg:proposal}).
The experiment is conducted with adversarial text generated by \textit{PWWS} targeting a CNN model on IMDB.
To eliminate the impact of support models,  we evaluate $\mathrm{VoteTRANS_{same}}$ without support as shown in Figure~\ref{fig:word_importace}. 
We also compare between $\mathrm{VoteTRANS_{same}}$ with word importance score and that with random score. 
$\mathrm{VoteTRANS_{same}}$ with word importance score achieves better than that with random score across all $\alpha$, especially with small $\alpha$. 
It demonstrates the impact of word importance score in $\mathrm{VoteTRANS}$.

\section{$\mathbf{WDR}$ Training on Other Word-Based Attacks}
\label{sec:appendix:word_based_attacks_training}

\begin{table}[t]
    \centering
    \begin{tabular}{l|c c | c c}
        \multirow{2}{*}{\textbf{Attack}} & \multicolumn{2}{c}{$\mathbf{WDR}$} & \multicolumn{2}{|c}{$\mathbf{VoteTRANS}$}\\
         & \textbf{F1} & \textbf{Recall} & \textbf{F1} & \textbf{Recall}\\
         \hline
        \textit{TextFooler} & 95.4 & 95.2 & \textbf{96.5} & \textbf{98.0}\\
        \textit{IGA} & 94.1 & 95.2 & \textbf{95.7} & \textbf{97.2}\\
        \textit{BAE} & 88.7 & 82.0 & \textbf{96.7} & \textbf{99.6}\\
    \end{tabular}
    \caption{$\mathrm{WDR}$ training on other word-based attacks.}
    \label{tab:WDR_other_attacks}
\end{table}

We train $\mathrm{WDR}$ on other main word-based attacks including \textit{TextFooler}, \textit{IGA}, and \textit{BAE} to detect adversarial text generated by \textit{PWWS} as shown in Table~\ref{tab:WDR_other_attacks}.
The adversarial text in both testing and training data targets the CNN model on AG News.
Since $\mathrm{VoteTRANS}$ performs without training, we use the corresponding attack as an auxiliary.
$\mathrm{WDR}$ is more compatible with \textit{TextFooler} and \textit{IGA} than \textit{BAE}.
In contrast, $\mathrm{VoteTRANS}$ achieves similar performance across three attacks.

\section{$\mathbf{VoteTRANS_{diff}}$ without Support}
\label{sec:appendix:other_diff_without_support}

\begin{table}[t]
    \setlength\tabcolsep{3.0pt} 
    \centering
    \begin{tabular}{l|c c}
         \textbf{Method}& \textbf{F1} & \textbf{Recall}\\
         \hline
        $\mathrm{VoteTRANS_{diff}}$ (\textit{Checklist}) & 
        41.4 & 26.4\\
        $\mathrm{VoteTRANS_{diff}}$ (\textit{PSO}) & 88.1 & 81.6\\
        $\mathrm{VoteTRANS_{diff}}$ (\textit{A2T}) & 88.2 & 82.0\\
        $\mathrm{VoteTRANS_{diff}}$ (\textit{Faster-Alzantot}) & 90.0 & 84.8\\
        $\mathrm{VoteTRANS_{diff}}$ (\textit{IGA}) & 90.1 & 86.0\\
        $\mathrm{VoteTRANS_{diff}}$ (\textit{Kuleshov}) & 90.8 & 86.4\\
        $\mathrm{VoteTRANS_{diff}}$ (\textit{TextFooler}) & 91.1 & 88.0\\
        $\mathrm{VoteTRANS_{diff}}$ (\textit{Input-Reduction}) & 93.8 & 97.6\\
        $\mathrm{VoteTRANS_{diff}}$ (\textit{Pruthi}) & 94.3 & 96.8\\
        $\mathrm{VoteTRANS_{diff}}$ (\textit{TextBugger}) & 94.6 & 97.2\\
    \end{tabular}
    \caption{Other auxiliary attacks for $\mathrm{VoteTRANS_{diff}}$ without support.}
    \label{tab:other_diff}
\end{table}

Besides \textit{BAE} and \textit{DeepWordBug} as mentioned in Table~\ref{tab:ablation_studies}, we show other auxiliary attacks in Table~\ref{tab:other_diff}. 
$\mathrm{VoteTRANS_{diff}}$ efficiently detects adversarial text except with \textit{Checklist}, which generates independent adversarial text with any model and does not focus on the performance.

\section{$\mathbf{VoteTRANS_{same}}$ with Other Supports}
\label{sec:appendix:other_supports}
\begin{table*}[]

    \centering
    \begin{tabular}{l|c c}
         \textbf{Method}& \textbf{F1} & \textbf{Recall}\\
         \hline
        $\mathrm{VoteTRANS_{same}}$ (\textit{PWWS}) with BERT as support & 97.5 & 100.0\\
        $\mathrm{VoteTRANS_{same}}$ (\textit{PWWS}) with ALBERT  as support & 97.6 & 99.6\\
        $\mathrm{VoteTRANS_{same}}$ (\textit{PWWS}) with DistilBERT  as support & 97.5 & 99.6\\
        $\mathrm{VoteTRANS_{same}}$ (\textit{PWWS}) with BERT+ALBERT as support & 98.4 & 98.4\\
        $\mathrm{VoteTRANS_{same}}$ (\textit{PWWS}) with BERT+DistilBERT as support & 98.2 & 98.8\\
        $\mathrm{VoteTRANS_{same}}$ (\textit{PWWS}) with ALBERT+DistilBERT as support & 98.2 & 98.8\\
        $\mathrm{VoteTRANS_{same}}$ (\textit{PWWS}) with BERT+ALBERT+DistilBERT as support & 98.4 & 99.2\\
    \end{tabular}
    \caption{Other available support models from \mbox{TextAttack} for $\mathrm{VoteTRANS_{same}}$.}
    \label{tab:other_support}
\end{table*}

\begin{table*}[!ht]
    \centering
    \begin{tabular}{l | l|c c}
         \textbf{Dataset} & \textbf{Method}& \textbf{F1} & \textbf{Recall}\\
         \hline
        \multirow{6}{*}{AG News} & $\mathrm{VoteTRANS_{same}}$ (\textit{DeepWordBug}) without support & 94.3 & 89.2\\
                                        & $\mathrm{VoteTRANS_{same}}$ (\textit{Pruthi}) without support & 66.4 & 76.8\\
                                        & $\mathrm{VoteTRANS_{same}}$ (\textit{TextBugger}) without support& 92.7 & 93.6\\
                                        & $\mathrm{VoteTRANS_{same}}$ (\textit{DeepWordBug}) with RoBERTa as support & 95.0 & 98.8\\
                                        & $\mathrm{VoteTRANS_{same}}$ (\textit{Pruthi}) with RoBERTa as support & 80.3 & 98.0\\
                                        & $\mathrm{VoteTRANS_{same}}$ (\textit{TextBugger}) with RoBERTa as support & 96.1 & 98.8\\                                        
        \hline
         \multirow{6}{*}{IMDB}& $\mathrm{VoteTRANS_{same}}$ (\textit{DeepWordBug}) without support & 90.5 & 93.2\\
                                        & $\mathrm{VoteTRANS_{same}}$ (\textit{Pruthi}) without support & 77.8 & 90.8\\
                                        & $\mathrm{VoteTRANS_{same}}$ (\textit{TextBugger}) without support& 92.1 & 95.2\\
                                        & $\mathrm{VoteTRANS_{same}}$ (\textit{DeepWordBug}) with RoBERTa as support & 93.3 & 99.6\\
                                        & $\mathrm{VoteTRANS_{same}}$ (\textit{Pruthi}) with RoBERTa as support & 82.0 & 98.4\\
                                        & $\mathrm{VoteTRANS_{same}}$ (\textit{TextBugger}) with RoBERTa as support & 97.6 & 99.6\\    
    \end{tabular}
    \caption{Detecting all character-based attacks compatible with CNN model from TextAttack.}
    \label{tab:character_attack}
\end{table*}

Besides LSTM and RoBERTa as mentioned in Table~\ref{tab:ablation_studies}, we show all other available support models for AG News from \mbox{TextAttack} and their combination in Table~\ref{tab:other_support}. 
VoteTRANS achieves performances of at least 97.5 with individual supports and their combination.
Similar to the results in Table~\ref{tab:ablation_studies}, multiple supports reach more stable results than an individual support.

\section{Detecting Character-Based Attacks}
\label{sec:appendix:char_attacks}

We conduct experiments to detect adversarial text from all character-based attacks from \mbox{TextAttack} compatible with the CNN model as shown in Table~\ref{tab:character_attack}. 
$\mathrm{VoteTRANS}$ achieves similar performances for AG News and IMDB. 
It demonstrates the resilience of $\mathrm{VoteTRANS}$ in detecting character-based attacks from different extents and tasks (medium text with multiclass classification as in AG News and long text with binary classification as in IMDB).

\section{$\mathbf{VoteTRANS}$ Complexity}
\label{sec:appendix:number_predictions}

\begin{figure}[]
    \begin{subfigure}[]{0.5\textwidth}
    \centering
    \includegraphics[width=0.95\textwidth]{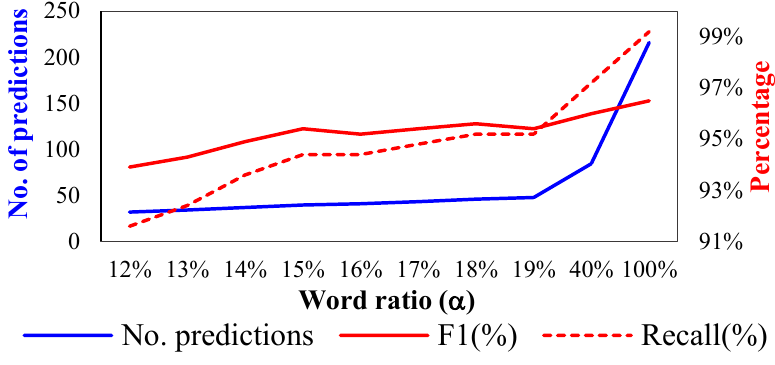}
    \caption{AG News.}
    \label{fig:number_of_prediction_agnews}
    \end{subfigure}
    \begin{subfigure}[b]{0.5\textwidth}
    \centering
    \includegraphics[width=0.95\textwidth]{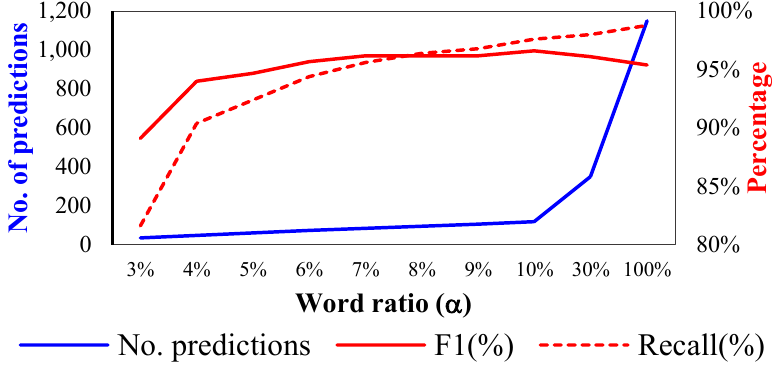}
    \caption{IMDB.}
    \label{fig:number_of_prediction_imdb}
    \end{subfigure}

    \centering
    \caption{Number of predictions.}
    \label{fig:number_of_prediction}
\end{figure}

Let $N$, $M$, and $K$ be the number of words, the number of transformations for each word, and the number of models used in Algorithm~\ref{alg:proposal}.
The worst-case of $\mathrm{VoteTRANS}$ complexity is $\mathcal{O}(N \times M \times K)$, approximately with the number of predictions on the $K$ models.
For example, if $\mathrm{VoteTRANS}$ processes AG News using \textit{PWWS}, CNN, and RoBERTa as auxiliary, target, and support; in this case, $N$, $M$, and $K$ are 42.6, 10.7, and 2, respectively.
$N$ of IMDB is increased to 241.9 while other values are unchanged.
Theoretically, the number of predictions is 910.6 (AG News) and 5165.6 (IMDB).
However, this number is remarkably reduced by the constraint checking (line 12) and early stopping (line 21) in Algorithm~\ref{alg:proposal}. 
As a result, it is reduced to 216.3 (76.2\%) and 1151.3 (77.7\%) predictions as shown in Figures~\ref{fig:number_of_prediction_agnews} and ~\ref{fig:number_of_prediction_imdb}, respectively.
$\mathrm{VoteTRANS}$ can also adjust the number of predictions suitable for the resource capacity by using small $\alpha$.
For example, $\alpha$ at 12\% and 5\% needs 32.0 and 61.1 predictions while keeping higher performance on the existing works as mentioned in Section~\ref{sec:run_time}.

\section{Detection with High Confidence Text}
\label{sec:appendix:high_confidence_adv}

\begin{figure}[!ht]
    \begin{subfigure}[b]{0.5\textwidth}
    \centering
    \includegraphics[]{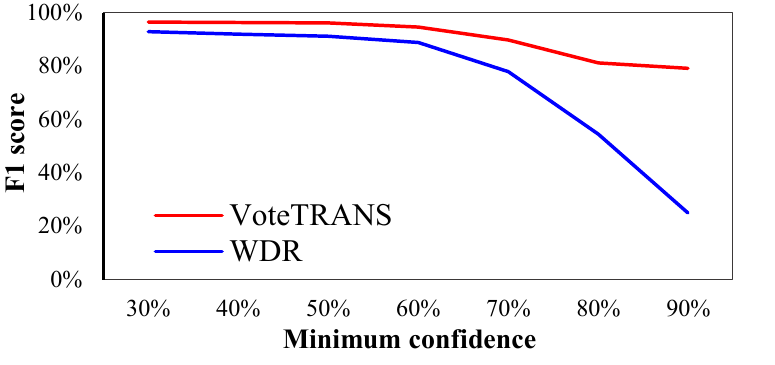}
    \caption{AG News.}
    \label{fig:high_confident_adv_agnew}
    \end{subfigure}
    
    \begin{subfigure}[b]{0.5\textwidth}
    \centering
    \includegraphics[]{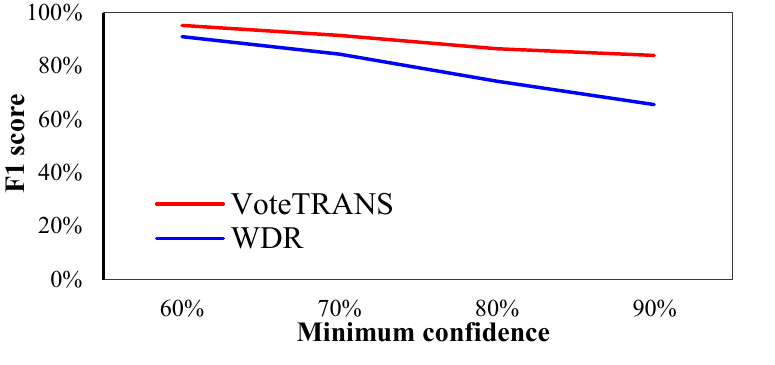}
    \caption{IMDB.}
    \label{fig:high_confident_adv_imdb}
    \end{subfigure}

    \centering
    \caption{Detection of adversarial text with high confidence}
    \label{fig:high_confident_adv}
\end{figure}

We evaluate $\mathrm{WDR}$ and $\mathrm{VoteTRANS}$ on detecting high confidence of adversarial text, which is generated by \textit{PWWS} targeting CNN models on AG News and IMDB.
PWWS attacks the CNN model until overcoming minimum confidence. 
Since any confidence of AG News (4 classes) and IMDB (2 classes) is greater than 25\% and 50\%, respectively, we set minimum confidences starting at 30\% and 60\% with 10\% step.
While the minimum confidences at 80\% and 90\% on AG News have 71 and 19 adversarial texts, respectively, other confidences have sufficient 500 balanced samples.

While $\mathrm{WDR}$ and $\mathrm{VoteTRANS}$ achieve similar F1 on AG News until minimum confidence at 60\%, $\mathrm{WDR}$ suddenly drops down to 25.0\% at confidence 90\%. 
In contrast, $\mathrm{VoteTRANS}$ still keeps 79.2\% at this confidence as shown in Figure~\ref{fig:high_confident_adv_agnew}.
For IMDB, the margins between $\mathrm{WDR}$ and $\mathrm{VoteTRANS}$ gradually increase from 4.2\% to 18.4\%.
It demonstrates the resilience of $\mathrm{VoteTRANS}$ in detecting adversarial text with high confidence.

\section{Error Analysis}
\label{sect_error_analysis}
We analyze the errors of $\mathrm{VoteTRANS_{same}}$ for the short text (MR). Here we especially focus on the results when DistilBERT and RoBERTa are the target and support models and the adversarial text is generated with \textit{PWWS}.
MR is harder to detect than long text as shown in Table~\ref{tab:comparison}. 
40.7\% of all the errors that $\mathrm{VoteTRANS_{same}}$ fails to detect are caused by susceptible original text, which is easily attacked. 
For example, although the original text ``\textit{your children will be occupied for 72 minute}'' is correctly predicted as negative by DistilBERT, 29 out of 39 perturbations with one-word replacements change the prediction into positive. 
It is opposite to our hypothesis as mentioned in line 38 in Section~\ref{sec:introduction} and thus bypasses $\mathrm{VoteTRANS}$. 
Its adversarial text ``\textit{your \underline{child} will be occupied for 72 minutes}'' also bypasses our detector (1 out of 42 perturbations change the prediction). 
However, such text is a little harmful because it has unclear sentiment and is unpopular (4.4\% and 1.2\% of MR and IMDB testing text, respectively).

\end{document}